\documentclass[runningheads]{llncs}
\usepackage[T1]{fontenc}
\usepackage{graphicx}
\usepackage{subcaption}
\usepackage{amsmath}
\usepackage{hyperref}
\usepackage{booktabs}
\usepackage{cite}

\hypersetup{
    colorlinks=false,
    pdfborder={0 0 0},
}
\begin{document}
\title{Model-based Clustering of Individuals' Ecological Momentary Assessment Time-series
Data for Improving Forecasting Performance}
\titlerunning{Model-based Clustering of Individuals' EMA Data}
%
\author{Mandani Ntekouli\inst{1} \and
Gerasimos Spanakis\inst{1} \and
Lourens Waldorp\inst{2}
\and
Anne Roefs\inst{3}
}
\authorrunning{M. Ntekouli et al.}
%
\institute{Department of Advanced Computing Sciences, Maastricht University, Maastricht, Netherlands \\
\email{\{m.ntekouli, jerry.spanakis\}@maastrichtuniversity.nl} \and
Department of Psychological Methods, University of Amsterdam, Amsterdam, Netherlands
\email{L.J.Waldorp@uva.nl} \and
Faculty of Psychology and Neuroscience, Maastricht University, Maastricht, Netherlands\\
\email{anne.roefs@maastrichtuniversity.nl}}
\maketitle              
\begin{abstract}
Through Ecological Momentary Assessment (EMA) studies, a number of time-series data is collected across multiple individuals, continuously monitoring various items of emotional behaviour. Such complex data is commonly analyzed in an individual level, using personalized models. However, it is believed that additional information of similar individuals is likely to enhance these models leading to better individuals’ description. Thus, clustering is investigated with an aim to group together the most similar individuals, and subsequently use this information in group-based models in order to improve individuals’ predictive performance. More specifically, two model-based clustering approaches are examined, where the first is using model-extracted parameters of personalized models, whereas the second is optimized on the model-based forecasting performance. Both methods are then analyzed using intrinsic clustering evaluation measures (e.g. Silhouette coefficients) as well as the performance of a downstream forecasting scheme, where each forecasting group-model is devoted to describe all individuals belonging to one cluster. Among these, clustering based on performance shows the best results, in terms of all examined evaluation measures. As another level of evaluation, those group-models' performance is compared to three baseline scenarios, the personalized, the all-in-one group and the random group-based concept. According to this comparison, the superiority of clustering-based methods is again confirmed, indicating that the utilization of group-based information could be effectively enhance the overall performance of all individuals' data.

\keywords{clustering \and  time-series \and  Ecological Momentary Assessment (EMA) \and  k-means \and  global forecasting models \and  Explainable Boosting Machine (EBM).}
\end{abstract}

\section{Introduction}
Using Ecological Momentary Assessment (EMA) methodology, multiple repeated measurements on an individual’s well-being and mood status can be easily collected, in respect to time and context information. This constitutes very important information in order to better understand people suffering from psychological-related issues and mental disorders \cite{nsmd}. Especially, nowadays, advancements in technology using smartphones and sensors have offered new opportunities in collecting more and more data for longer time periods for each individual, eventually organized as multivariate time-series data. However, data are not always easily provided. Due to several factors, such as lengthy collection periods or sampling frequency, a number of collected measurements can still be missing \cite{missing}. This leads to the fact that too little data are not well-representative of a user’s behavior and consequently these are sometimes not enough to train accurate and reliable personalized models. 

A way to approach this issue could be to provide information collected by other similar individuals during the same EMA study. Although, it is known that every individual is unique and exhibits their own symptoms and behaviors, it is likely that shared patterns can also be found in group of people. As a result, using additional information of similar individuals, sharing common driven forces and behaviors, could be very useful for the majority of them. It could be used for better describing and understanding the profiles of single or groups of individuals, uncovering hidden structures as well as building more accurate prediction models for forecasting future behaviors.

Finding similar patterns among objects, or individuals in the current problem, when the true grouping is not available, is known as a clustering task. So, clustering methods could be utilized on EMA data, with the goal of finding the optimal groups of similar individuals \cite{ntekouli2022clustering}. Clustering has been studied a lot, with a great interest to time-series data in the recent years \cite{decade, rev1}. Although most straightforward and popular clustering approaches use raw time-series data and further researching the most appropriate similarity/distance measure, other type of representational information can be also used to characterize each individual. For example, model-derived information, such as model's coefficients, could be also utilized, reflecting another promising clustering category, that is model-based clustering \cite{k-model}.

In model-based clustering, the goal is to find similar groups of models that possibly represent different groups of individuals. In other words, each individual is described by a model, and, more specifically, several model’s parameters. In this case, the raw time-series data are used as well, but to build the prediction models, not directly for clustering. Then, clustering is applied on the model-derived information of each individual. This information may include different characteristics of each model. For instance, for linear models, it can be the extracted coefficients of the trained models. Thus, identifying similar sets of coefficients could be useful for uncovering similar individuals.

This paper aims to investigate the use of model-based characteristics or information for clustering high-dimensional time-series EMA data, through two different approaches.
First, model-derived parameters, like coefficients or feature importance, of individual models are exploited for applying clustering \cite{k-model}. Second, since one of the clustering goals is to improve the forecasting performance, performance could be also considered as an alternative information used to optimize clustering \cite{perf}. 
To evaluate both clustering approaches, all clustering scenarios are assessed regarding some intrinsic evaluation measures, such as silhouette coefficients and stability, as well as post-hoc forecasting performance. In terms of performance evaluation, clustering methods are also compared to three baseline scenarios, the personalized, the all-in-one group and the random group-based approaches \cite{ntekouli2022using, idio-nomo}.

\section{Related Work}

In this section, first, related work to raw time-series clustering will be described followed by model-based clustering approaches for time-series, which is the emphasis of this work.

\subsection{Time-series Clustering}
Time-series clustering has been studied a lot lately, with some handful reviews found in \cite{rev1, decade}. According to these, the majority of works focus on clustering based on the raw time-series data while exploring different choices for clustering method, distance metric and evaluation. In such an approach, the most important decision is definitely the chosen time-series distance/similarity metric that will be later used in all well-known clustering methods, like k-means or hierarchical clustering \cite{ntekouli2022clustering}. Most of distance measures may be based on the concept of intensity distance or shape resemblance \cite{kml-shape}. While the majority of studies consider two individuals similar if their variables intensity at each time point are close, this does not take into account shape information. Thus, two same variable shifted in time cannot be considered similar. To deal with such common issues in time-series, recently, shape-based distance metrics, such as dynamic time warping (dtw), are widely applied, trying first to optimally align the data. Because of their success in many time-series applications, different dtw-related metrics have been also developed \cite{app_dtw}. Some examples are soft-dtw, the global alignment kernel similarity and the longest common sub-sequence \cite{soft}. Alternatively, in a less expensive computational manner, cross-correlation has also been applied as a distance metric for clustering time-series in \cite{k-shape}. As a next step, clustering is applied through incorporating or adopting these applied distance measures within a clustering method, and subsequently followed by some evaluation measures.

\subsection{Model-based Clustering}
Apart from using the raw time-series, different data representations can also play a key role when applying clustering \cite{rev1, decade}. These include different statistical-derived data features, data reduction and transformation techniques, like Principle Component Analysis (PCA)-representation, with all trying to differently capture time-series dynamics. This step is then followed by the clustering method. Similarly, as an alternative time-series representation approach, model-derived features and characteristics could be also used. This is considered as the model-based clustering category.  

According to two review papers on time-series clustering, there are basically two main model-based categories \cite{rev1, decade}. The first approach starts with the assumption that each individual’s data can be reliably described by a model, so that it can be represented by the model’s estimated parameters. As a result, the problem of finding the most similar individuals is translated into finding the most similar parameters among the different models. In this case, different types of parameters can be used, depending on the applied base model. For example, parameters can be easily extracted from linear models, such as Auto-regressive (AR) model or Auto-regressive Integrated Moving Average (ARIMA) \cite{ar-metric}, where the auto-regression coefficients are then used. Various transformations of these can be also be exploited by clustering \cite{ar-tr}. Similarly, using probabilistic models, like Hidden Markov Model (HMM) \cite{hmm}, clustering is applied to the probability densities derived from individual HMM. In the case of non-linear models, other type of parameters can be produced. For instance, for Random Forest (RF), measures relying on variables importance values have been proposed \cite{k-rf, rf-imp}.

Recently, more complex model have been also studied \cite{deep}. In \cite{rnn}, the parameters derived from the output layer of an Recurrent Neural Network (RNN) model are utilized as the input for clustering. These model-based dynamic parameters incorporate more complex information, capturing both the temporal dynamics and the local structures of the data.

On the other hand, the second category aims at recovering the optimal data partition that fits a mixture of group models that better represent the whole set of individuals. More specifically, it is alleged that each individual is optimally described by a group model trained on a set of similar individuals, individuals belonging to the same cluster. In such settings, each group model would correspond to one cluster and the data of each individual to the input data used to train the group models. Commonly, the models used in this approach can be based on statistical and probabilistic methods. This category includes methods estimated by the Expectation-Maximization (EM) algorithm, such as Gaussian Mixture Models (GMM) and HMM. Using EM, the produced mixture of models always results to a soft-clustering solution. Additionally, hard-clustering is possible if, instead of EM, a version of k-means is used. Such an approach was recently studied in \cite{k-model}. The proposed K-Models paradigm aims to fit a separate model for each cluster, tested on AR, ARMA and ARIMA. 

However, it is worth highlighting that all these approaches are not easily adapted to multiple multivariate time-series data of unequal lengths. This is a challenge that will be explored in this work, where we have a clustering problem of multivariate time-series of multiple individuals. Moreover, another goal that the current paper explores is the application of both aforementioned approaches with more advanced non-linear models.

\section{Methodology}
This section starts by describing the structure of the examined EMA time-series data. Then, some more details are provided regarding the forecasting models that play a key role in the proposed clustering procedure. Finally, we present thoroughly the two model-based clustering approaches we focus on this work. While the first one is based on the parameters of already trained individual-forecasting models, the second is focusing on training group models as part of the clustering procedure with the goal to find the most representative group models for all individuals.

\subsection{EMA Data}
The data used for testing our approach follow the typical structure of EMA time-series data. 
In more detail, $X$ denotes the entire EMA dataset, including the sub-data of all $N$ individuals, $X =  \text{\{} X_1,.., X_N \text{\}}$. Each individual data $X_i$, where $i$ ranges from 1 to $N$, is then described by $V$ variables repeated over $T_i$ time-points. So, $X_i$ is multivariate time-series dataset, defined as $X_i^{1..V, 1..T_i}$. While the $V$ variables are the same over the entire dataset $X$, the time-points collected for different individuals can differ. This is common for EMA data and caused by cases like missing or unanswered questions \cite{missing}.

\subsection{Introduction to Forecasting Models}
\label{forecast}
Following, we establish one of the main procedures in the examined approaches, besides clustering, that is regarding the training of forecasting models. It is important to highlight the key role of the used forecasting models, since the whole clustering procedure is relying on these.

In general, the task of forecasting models aims at accurately predicting the future responses. However, in the current problem, the task becomes more complicated by aiming to the 1-lag future values of all variables. In a multivariate time-series setting with $V$ variables, that means that we need to build $V$ independent models, each using the same input predictors, the set of all variables, and predicting one of the examined variables shifted in the future. Thus, when referring to one individual forecasting models, this implies all $V$ independent sub-models necessary for predicting all $V$ variables in the future.

\subsection{Model-based Clustering Approaches}
\label{approaches}
The two examined model-based clustering approaches are presented. 

\subsubsection{Approach I: Parameter-Driven Clustering (PDC)}
\begin{figure}
    \centering
    \includegraphics[scale=0.25]{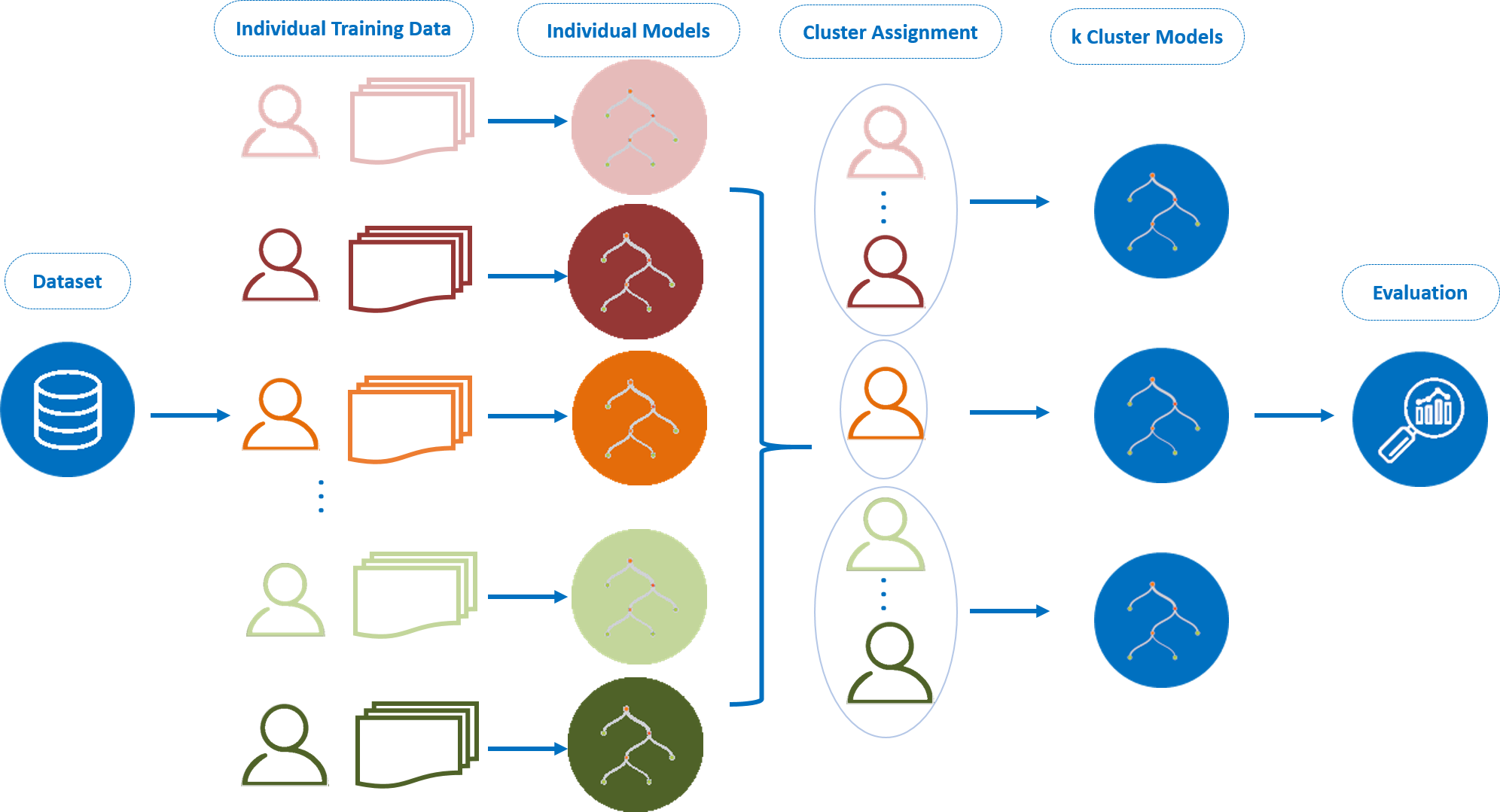}
    \caption{Clustering approach I: Parameter-Driven Clustering (PDC). This approach considers the parameters extracted by the $N$ individual models as the input of clustering.}
    \label{fig:ap1}
\end{figure}

According to PDC, clustering is performed using model-derived information representing each individual. In particular, this approach considers the parameters extracted by the $N$ individual models as the input of clustering, as presented in Figure \ref{fig:ap1}. In other words, it uses the parameters of trained models to represent every singe individual, assuming that these models can accurately describe them. For each individual, $V$ variables need to be predicted, which, according to the EMA structure where a variable can be predicted by values of all variables in previous time-step, is possible through $V$ different models.
So, the parameters of all $V$ independent models need to be concatenated in order to better represent each individual.

The type of the parameters depends on the base individual model that is used. Here, different base models are used, both linear and non-linear ones. On the one hand, in case of linear models, the fitted coefficients can be easily extracted. Each coefficient indicates the influence of one variable in predicting the future values of another. So, to cover all combinations, finally $V \times V$ coefficients are used for representing one individual. Finally, the coefficients' matrix, that is the set of $N$ by $V \times V$ parameters, is input to clustering.

On the other hand, for non-linear models, such coefficients are not inherently available. Thus, in a similar way, trying to quantify the influence of one variable to another, feature importance values can be used. In this scenario, the number of parameters for each individual has remained the same, $V \times V$, as well as the total parameters matrix, $N$ by $V \times V$. 

\subsubsection{Approach II: Performance-Optimized Clustering (POC)}
In the second approach, clustering is applied using a different model-derived information, predictive performance, that is assumed to alternatively describe the individuals. In this case, clustering aims at building representative global or cluster models, each consisting of similar individuals based on their forecasting performance \cite{perf}. Each cluster is basically optimized on the total test performance of all individuals, in terms of the Mean Squared Error (MSE) across all variables and time-points of their test set. 

In particular, the procedure is quite similar to the original k-means algorithm, with the main difference found in the objective function. The goal now is to minimize the MSE errors on the test set of all individuals, instead of minimizing the within cluster distance of all individuals to the centroid. In more details, the different steps are summarized in Figure \ref{fig:ap2}, and described as follows:
of are the following:

\begin{figure}
    \centering
    \includegraphics[scale=0.25]{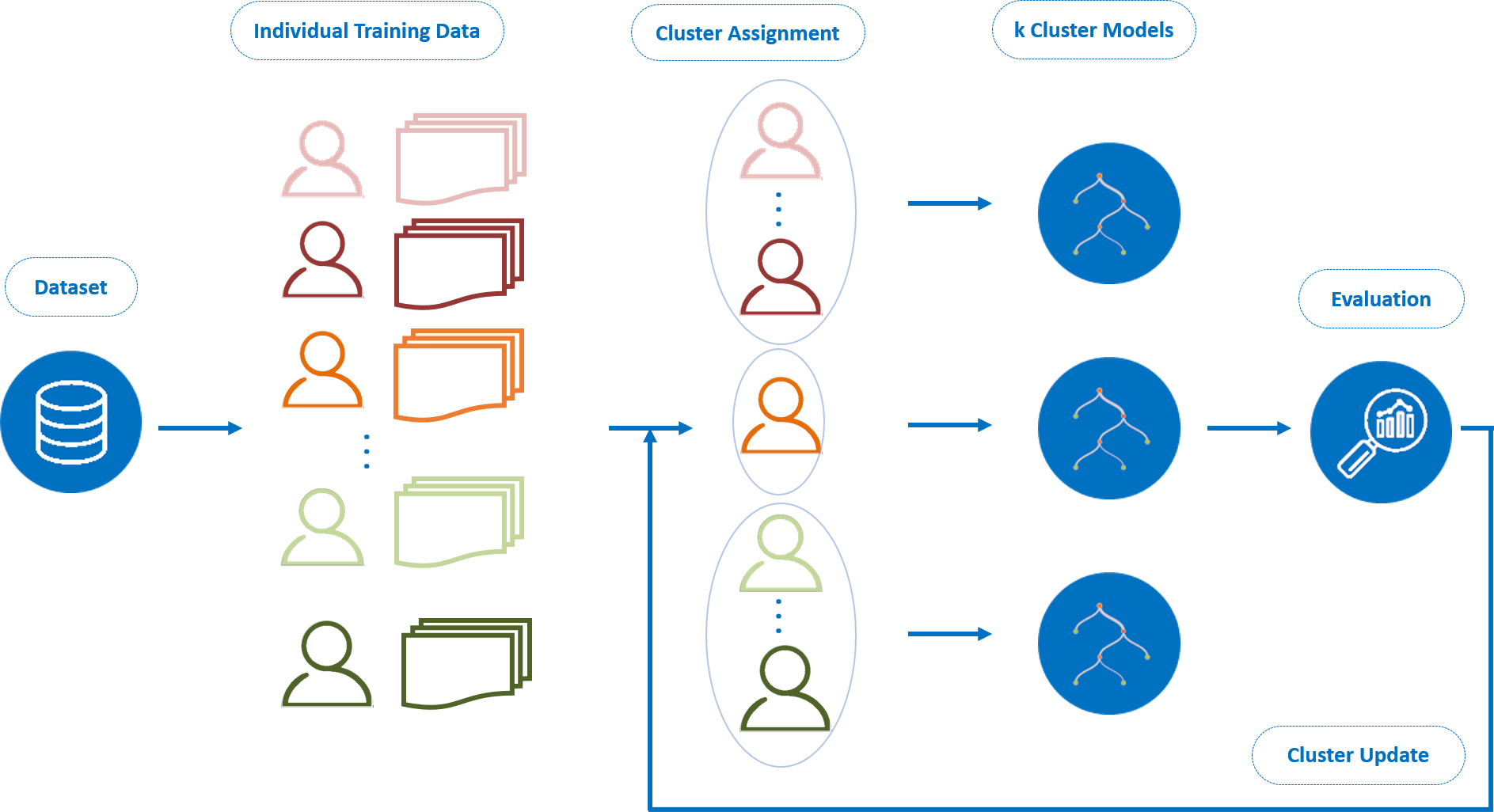}
    \caption{Clustering approach II: Performance-Optimized Clustering (POC). This approach is optimized on forecasting performance, ultimately aiming at building $k$ representative cluster models.}
    \label{fig:ap2}
\end{figure}

\begin{itemize}
    \item Initialization Step: The procedure starts with a random initialization, setting the centroids of the clusters. For faster convergence, we follow an approach similar to k-means++ algorithm. More specifically, after the first centroid is randomly selected, the rest should be set in a way to be the most dissimilar to the first one. To find the most dissimilar individual to the centroid, it is assumed that its performance on a model based on this individual-centroid should be the worst. So, we start by building a model associated to the first centroid (centroid model) using the data of the first selected individual-centroid. Then, this model is tested on all the rest $N-1$ individuals and the one with the highest test error is then selected as the individual-centroid of the another cluster. If the predefined number of clusters ($k$) is greater than 2, this the process is repeated until the centroids of all clusters have been found.
    \item Clusters Assignment Step: To assign the rest of the $N-k$ individuals to a cluster, again their performance is used as a measure. Instead of calculating the actual distance to the clusters’ centroids, their performance is examined, in terms of MSE error, on the $k$ centroid models. As a result, each individual is assigned to the cluster with the minimum error.
    \item Forward Step: Having assigned all individuals into clusters, the main procedure begins with the goal to optimize the objective function of the predictive algorithm. This is described in \ref{eq:loss}, where the total test MSE error $L_{MSE}$ corresponding to all $N$ individuals need to be minimized, meaning the error of every single individual. To achieve this, the cluster models are built using the training data (the first part of dataset) of the individuals belonging to each cluster. Then, all the cluster models are assessed on the test set (the last part of each dataset, where time-points ranging from 1 to $T$) of all individuals, predicting all $V$ variables.

\begin{equation}
\label{eq:loss}
    L_{MSE} = \sum_{i=1}^N \frac{\sum_{t=1}^{T} \sum_{v=1}^{V} (x_i^{v,t} - \hat{x}_i^{v,t})^2}{T\cdot V}
\end{equation}
    
    \item Update Step: For each individual, its test performance is compared on all the cluster models. If the test MSE error is smaller in another cluster than the one already belonging to, the individual move to the other cluster. This way, all the clusters composition is updated. 
\end{itemize}

The last two steps, forward and update, are repeated until there is a convergence, meaning that individuals are fixed in clusters, or the maximum number of iterations have been reached. When the clustering algorithm stops, the clusters composition have been finalized, the group models are trained and the total minimum loss has been found. However, beyond the initialization step, clusters centroids are not directly used or calculated.

\section{Experiments}
Through the following section, the results of the two model-based approaches are presented. First, both are tested using different base models, such as Explainable Boosting Machine (EBM) \cite{interpret-ml}, where clustering and final forecasting performance are examined. Then, the final performance of the clustering approaches is compared against some baseline approaches. These include the personalized, also referred as N-clustering problem, and all-in-one group, or 1-clustering, forecasting models as well as group models where individuals were randomly assigned.

\subsection{EMA Time-series Data}
The examined data is a real-world EMA dataset obtained by a student pilot study in the Netherlands, aiming at assessing the questions and responses collected for the real study, which is planned to target people with mental disorders \cite{martinez2023developing}. The dataset initially included information of $277$ individuals, monitored 8 times a day for 4 weeks. This leads to a maximum of 224 time-points per individual. However, during the pre-processing step, after removing people with less than 50\% compliance (112 time-points), 187 individuals finally remained. Their missing points were deleted after creating the next time-point labels for each variable, so it didn't affect the training process. Then, for each individual, 37 variables were collected in every questionnaire, mostly referring to emotions and context. According to domain-experts knowledge, these variables were decreased to 12, after merging some similar ones and removing the ones with low variance or information.

\subsection{Experimental Setup}
In the analysis, the following set of experiments are investigated. To assess the clustering performance, three baseline scenarios also examined are the N-clustering (or personalized), the 1-Clustering (all individuals belong to one group) and random-groups clustering, as follows:

\begin{itemize}
    \item \underline{k-Clustering}: During k-means clustering, the number of clusters ($k$) needs to be predefined. Here, the values of $k$ is set in a range from 2 to 20. When $k$ is set, we compare the results with a different base forecasting model, where here there is a set of 3 different models. The set includes one linear, the Vector Auto-regressive (VAR) model, and 2 non-linear ones, Random Forest (RF) and Explainable Boosting Machine (EBM) \cite{interpret-ml}. For each combination of $k$ and forecasting model, k-means clustering is then repeated for 10 iterations.

    \item \underline{N-clustering}: The total forecasting performance after clustering is compared with the case of having personalized models. So, for each individual a separate model is trained based on its own data and assessed on its unseen test data.  Having $N$ individual models for $N$ individuals is identical to the case clustering using $k=N$. Thus, the concept of using personalized models is also called N-clustering. 

    \item \underline{1-Clustering}: In a similar manner, another baseline scenario under investigation involves the case of $k=1$, which is also referred as 1-Clustering. In this case, it is assumed that all individuals belong to 1 cluster, so that only one model could describe them all after being trained on all data. For a fair comparison, its performance is tested separately on the test data of each individual.
    \item \underline{Random-Clustering}: To ensure that the effect on clustering performance is not caused just from the fact that less individuals than all, but also more than one, are used in a model, some additional experiments are added. The case of randomly assigning individuals to different clusters is then examined. The same range of $k$ values is again used and 10 iterations are executed for each experiment.
\end{itemize}

\subsection{Evaluation}
After applying clustering, there are several ways to evaluate the derived results. Here, the evaluation is conducted through the quality of clusters (intrinsic measures) as well as the performance of a downstream forecasting model.

\begin{itemize}
\item \underline{Clustering Evaluation}:
\label{eval}
For clustering evaluation, either intrinsic or extrinsic measures \cite{survey} are mostly used. Because the latter one requires obtaining ground truth labels, which are not available in our case, the focus of this work is on the intrinsic measures. The majority of the intrinsic measures are computed using 
the compactness of each cluster as well as how well different clusters are separated. In our analysis, the well-applied Silhouette coefficient is selected, since other measures (such as Davies-Boudlin \cite{survey}) are mostly based on clusters' centroids, which lacks of any natural explanation. Silhouette coefficients are given through the equation \ref{eq_sil}, calculating the average score across all $N$ individuals. For each individual $i$ belonging to cluster $C_i$, the score compares the average similarity $a_i$ across all individuals of the same cluster ($w$, where $w=1..W_i$ and $W_i$ is the total number of individuals of $C_i$) to the average similarity $b_i$ of the individuals belonging to the closest to $C_i$ cluster ($z$, where $z = 1..Z_i$ and $Z_i$ is the total number of individuals of the closest to $C_i$ cluster).

\begin{align}
\label{eq_sil}
    & Sil = \displaystyle \frac{\sum_i^N\frac{ b_i-a_i}{ \max(b_i,a_i)}}{N} , \text{where }\\
    &  a_i =  \frac{\sum_{w \epsilon C_i} d(x_i, x_w)}{W_i} \text{ and } b_i = \frac{\sum_{z \epsilon \text{ closest } C_i} d(x_i, x_z)}{Z_i} \nonumber
\end{align}

Another evaluation measure is the stability of the clustering results \cite{stab}. Through repeating clustering multiple times, the stability or agreement of individuals assignment into clusters can be assessed. More specifically, the Adjusted Mutual Information (AMI) between the individuals' cluster labels is used as stability index \cite{adjusted}.

\item \underline{Forecasting Performance Evaluation}:

An additional way to evaluate clustering performance is through the performance of a downstream forecasting model. Using the data of each of the clustering-derived groups, different cluster model can be trained aiming at forecasting the 1-lag future values of all variables, as described in Section \ref{forecast}. Then, these can be evaluated using the mean squared error (MSE) on the test set of each individual. Particularly, each individual is evaluated on the MSE of their test set data (last 30\% of its data) using the cluster model that belongs to. The total MSE of all individuals on their test sets is used as the performance indicator of each clustering method. Similarly, the performance of the baseline methods can be assessed, using the total MSE on the same test sets of all individuals.

\end{itemize}

\subsection{Results}
The results of the conducted experiments are presented in this section. First, the two proposed clustering methods, PDC and POC, are assessed through different evaluation measures. 
Afterwards, clustering is assessed through its downstream forecasting performance as well as in comparison to three baseline methods, 1-clustering, N-clustering and Random-clustering.

\subsubsection{Clustering Evaluation}
As already described in Section \ref{approaches}, for cluster analysis, or, as we also call it, k-Clustering, two model-based approaches are investigated, based on coefficients (PDC) and performance (POC), respectively.
In both approaches, different choices are examined, such as the applied base models, which can be VAR, RF and EBM, while the number of clusters ($k$) can also take values from the set $\{2, 3, 4, 5, 6, 10, 15, 20\}$. Regarding PDC, the selected base model refers to both individual and cluster models. Then, for both approaches, all combinations of these choices represent the examined experiments, where each one is repeated 10 times.

\begin{figure}[t]
     \centering
     \begin{subfigure}[t]{0.45\textwidth}
         \centering
         \includegraphics[width=\textwidth]{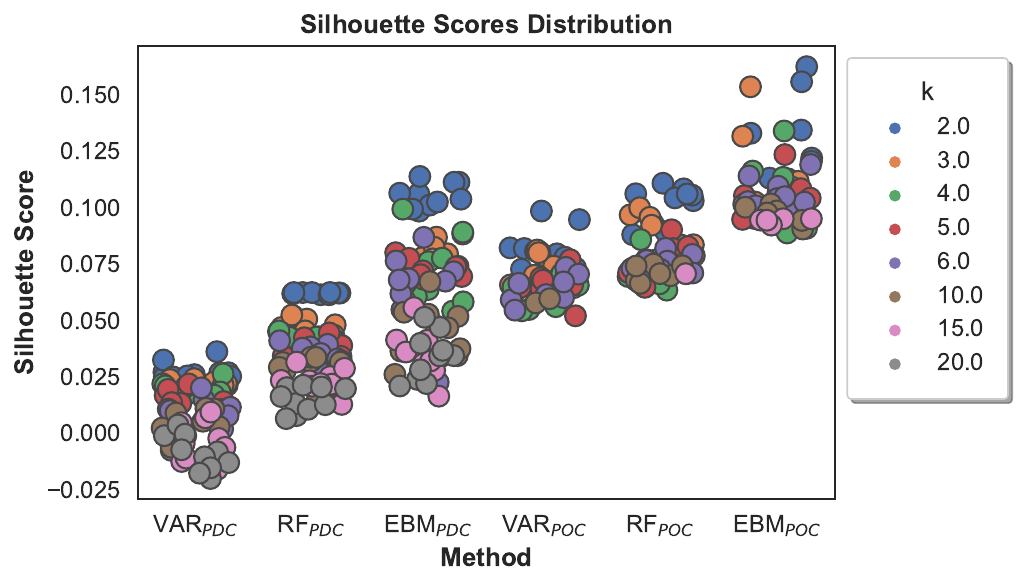}
         \caption{Based on the model-based parameters used.}
         \label{fig:sil1}
     \end{subfigure}
     \hfill
     \begin{subfigure}[t]{0.45\textwidth}
         \centering
         \includegraphics[width=\textwidth]{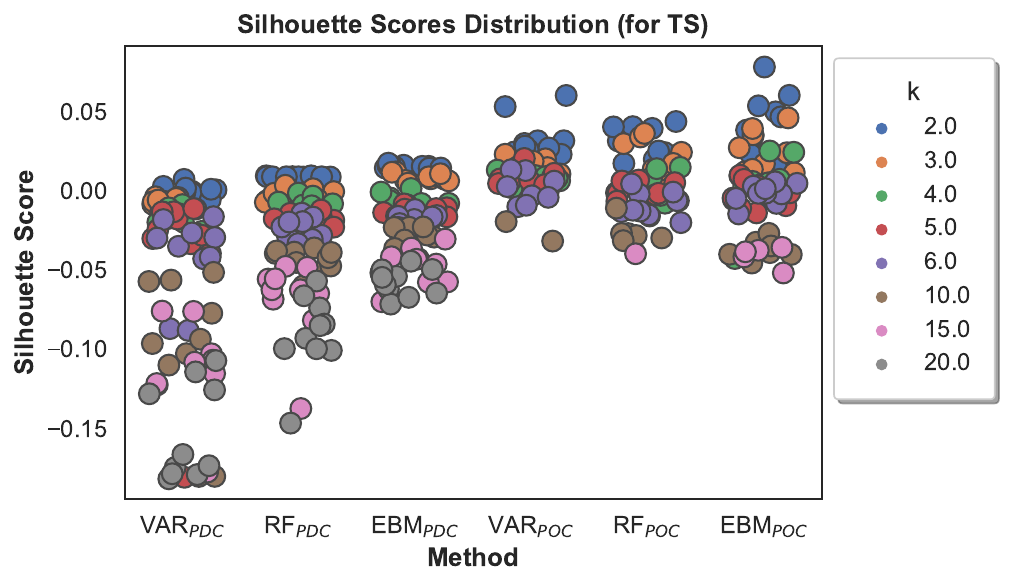}
         \caption{Based on the dtw distance of individuals time-series data.}
         \label{fig:sil2}
     \end{subfigure}
     \caption{The silhouette coefficients of all experiments of both approaches (PDC and POC) are presented. The difference between the chosen values of $k$ and base models is depicted. The same-coloured points represent different iterations of the same experiment.}
\end{figure}

All these experiments are evaluated using the two clustering intrinsic evaluation measures, described in Section \ref{eval}, Silhouette coefficients and Stability. 
First, the extracted Silhouette coefficients of all the experiments of both approaches are depicted in Figure \ref{fig:sil1}. For each method on the y-axis, all points of the same colour represent the 10 iterations each experiment was repeated using a particular value of $k$. Between the two clustering approaches, the second one, POC, clearly reaches much higher scores than the first one. This holds for all different base models. In detail, EBM gives the highest maximum scores in both approaches, around $0.17$ and $0.11$, respectively, whereas the VAR models shows the lowest maximum scores at $0.10$ and $0.03$, respectively. Although the second approach mostly outperforms the first one, when using EBM with low value for $k$, the difference is not that significant. Thus, EBM models seem to better describe the complex EMA data, even in the form of feature importance values. 

It is also noticed that increasing the number of clusters used, the produced Silhouette tend to decrease. Regarding the first approach, there are iterations for EBM indicating that $k=2$ and $k=4$ give the best clustering, whereas for the second approach, that occurs for $k=2$ and $k=3$.

In a similar way, according to the derived clustering grouping, the Silhouette coefficients calculated based on the dtw distance of the individual time-series data can also be calculated and shown in Figure \ref{fig:sil2}. The found pattern resembles the one of \ref{fig:sil1}, although the Silhouette values are lower for all the experiments. In the first approach, all coefficients are below zero, indicating a meaningless clustering. However, referring to the second approach, the scores for low values of $k$ are a bit higher, but not exceeding the value of $0.07$, when using $k=2$ and EBM. 

\begin{figure}[t]
     \centering
     \includegraphics[width=0.5\textwidth]{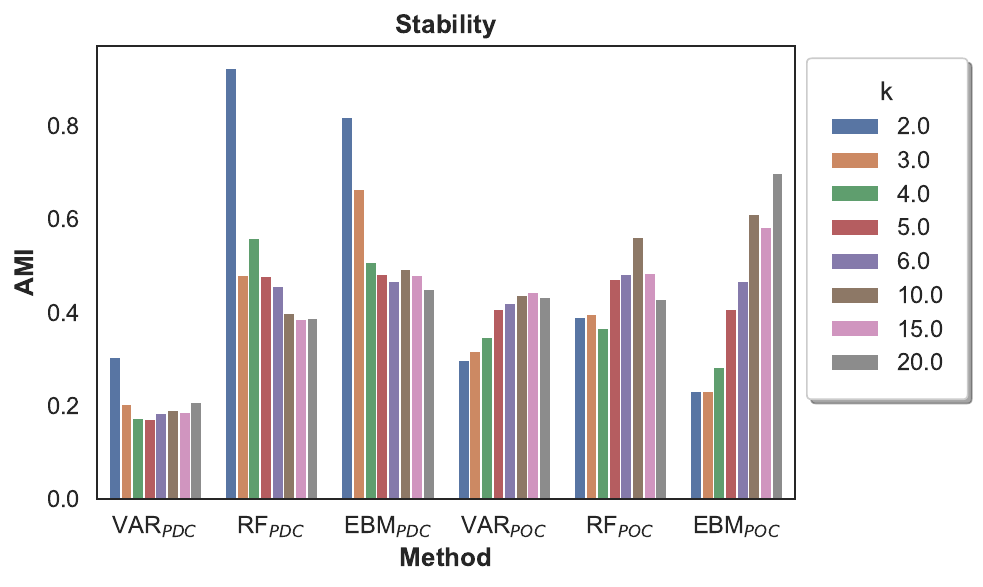}
     \caption{Stability of all experiments of both approaches (PDC, POC) is presented.}
     \label{fig:stab}
\end{figure}

Next, all clustering experiments of both approaches are assessed for their stability, in terms of cluster labels agreement across all iterations. The experiments' stability, as expressed by the AMI scores, is presented in Figure \ref{fig:stab}. In most of the cases, it is obvious that the stability values are quite low. This is caused by the fact that the produced groups were quite different to each other. Based on the complexity of individual EMA patterns, it seems unlikely for different people to be always separated in a particular way. 
Thus, we'd better work with the group information each iteration separately extracts. 

It is important to also note that, even though k-means requires a prior specification of the number of clusters (given $k$), the number of produced clusters could deviate from this. So, in the whole analysis, only the iterations of experiments that actually produce the given number of clusters are considered valid. While in PDC, clustering always produces the given number of clusters, for POC the number mostly deviates from the expected. The average number of clusters is displayed in Table \ref{tab:counts}. When the average produced number is smaller than the given one, it means that there are iterations with less extracted clusters. This is usually the case when increasing the chosen number of clusters. These are eventually excluded from the analysis of Silhouette coefficients and Stability.

\begin{table}[t]
    \centering
    \caption{The average number of clusters, across 10 iterations, produced by the second clustering approach, for all examined experiments.}
    \begin{tabular}{lrrrrrrrr}
k &  2  &  3  &  4  &  5  &  6  &  10 &  15 &  20 \\
\midrule
VAR &   2.0 &   3.0 &   4.0 &   4.8 &   5.8 &   9.2 &  12.6 &  13.6 \\
RF  &   2.0 &   3.0 &   4.0 &   5.0 &   5.9 &   9.7 &  13.6 &  17.2 \\
EBM &   2.0 &   3.0 &   4.0 &   5.0 &   6.0 &   9.7 &  14.4 &  18.1 \\
\bottomrule
\end{tabular}
    \label{tab:counts}
\end{table}


\subsubsection{Downstream Forecasting Performance}
\begin{figure}
     \centering
     \begin{subfigure}[t]{0.45\textwidth}
         \centering
         \includegraphics[width=\textwidth]{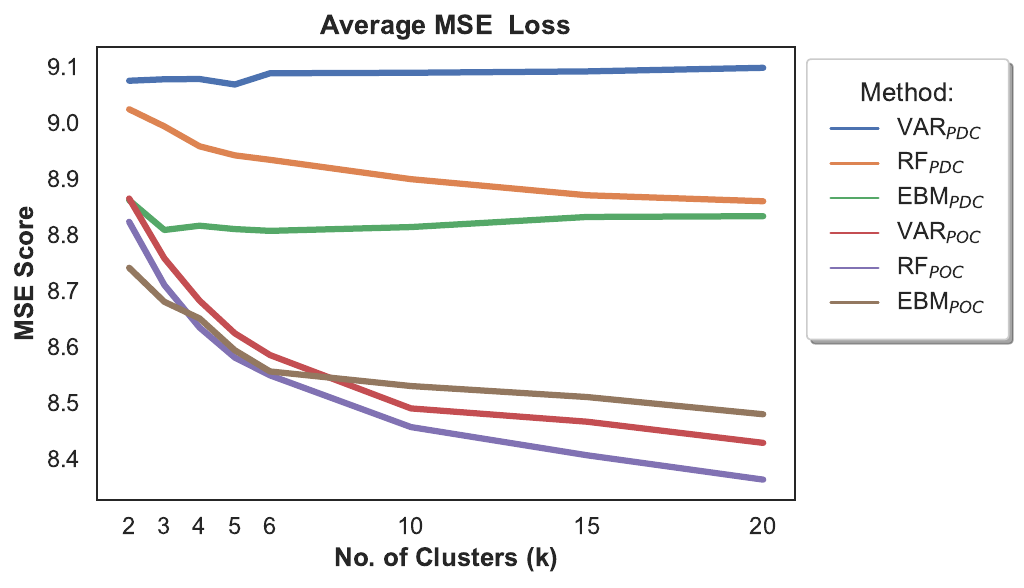}
         \caption{The experiments of the two proposed clustering approaches (PDC and POC) are compared.}
         \label{fig:loss}
     \end{subfigure}
     \hfill
     \begin{subfigure}[t]{0.45\textwidth}
         \centering
         \includegraphics[width=\textwidth]{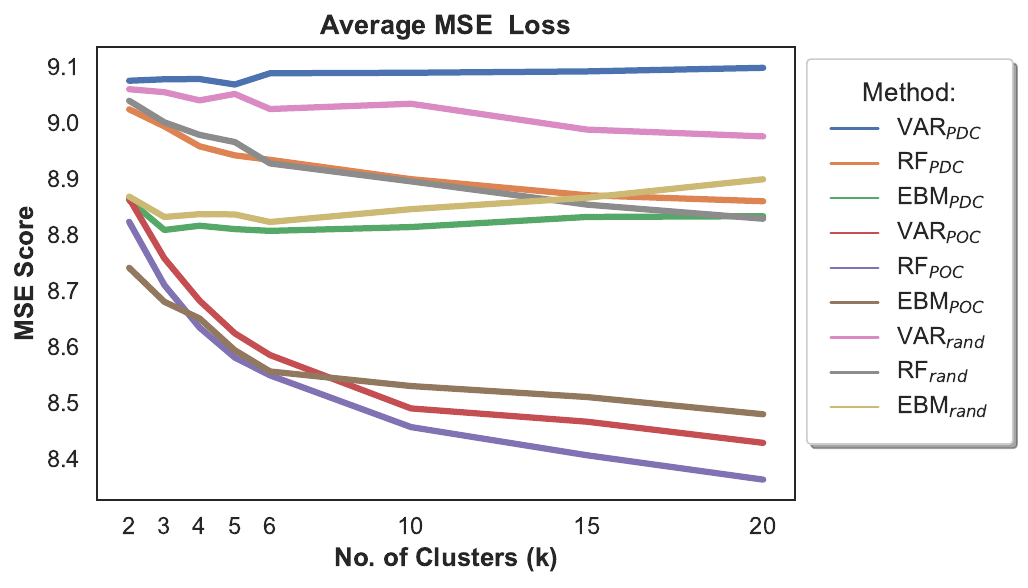}
         \caption{The two clustering approaches (PDC and POC) are compared to the random individuals grouping (rand).}
         \label{fig:loss2}
     \end{subfigure}
     \caption{The total MSE loss of all individuals is given, averaged across all iterations.}
\end{figure}

As a second step, clustering can be evaluated through the forecasting performance of the downstream clustering-derived group models. First, the total MSE loss scores of all individual test sets are compared across all experiments of the two proposed clustering approaches. The average loss scores of the 10 iterations of all experiments are shown in Figure \ref{fig:loss}. A clear distinction is again obvious between the two clustering approaches, with POC leading to slightly lower loss scores, that is translated to a better performance. 

In more detail, on the one hand, for the first approach, the scores do not vary a lot for the examined number of clusters, leading to a score approximately at $9.07$, $9.02$ and $8.86$ in the cases of VAR, RF and EBM models, respectively. On the other hand, for the second approach, the scores are much lower and not that constant across the different values of $k$. While for low values of $k$, the loss indicates that EBM show the best performance, this changes after $k=6$, where RF, then, gives the best scores. Although that difference between the base model is not important, the difference across the $k$ values is quite larger. For all base models, it starts at around $8.7-8.9$ and it ends at $8.3 - 8.5$. Thus, increasing the number of clusters, and consequently that of the cluster-models, seems to improve the performance. However, in that case, the number of the actual clusters found tends to be smaller than the given one. Therefore, there should be a trade-off between the performance improvement and the capability of finding the given number of clusters.

As a general result, the produced loss scores are partly in agreement to the findings after the Silhouette analysis. Both show that the POC outperforms PDC, while clustering using EBM, as the base model, yields mostly the best performance. Nevertheless, the impact of different $k$ values is not consistent. Although, the performance is optimal when $k=20$, the Silhouette coefficient shows its lowest value.

Because of this disagreement in the analysis, we should further reject the possibility of having enhanced performance results by just using more cluster models, instead of meaningful cluster models. In other words, if this improvement could be a random effect of just splitting the individuals in more clusters, so having more cluster-models in number. To do this, we split individuals randomly to the same predefined range of possible clusters, without having applied any clustering technique. Again, every experiment is conducted 10 times, and the average loss scores are exhibited in Figure \ref{fig:loss2}. It is observed that the loss line plots of random grouping follow the patterns of the first approach. Also, it is quite worse than the experiments of POC. Thus, it seems that the improved forecasting performance is not connected to a random effect of just using more cluster-models.

Finally, the two proposed clustering approaches are compared to two baseline scenarios, namely N-clustering and 1-Clustering, which basically correspond to the personalised and all-in-one group methods, respectively. For all these scenarios, the total loss scores of all individuals (averaged on all variables and test time-points for each individual) are given in Table \ref{tab:2}. As for k-Clustering, both $k=2$ and $k=20$ clusters are considered, which, in POC, yield better performance compared to the baseline scenarios. According to the derived scores, the lowest loss is found when using 20-Clustering and RF at $8.36$. Regarding the baseline scenarios, the lowest MSE score is for the personalized EBM at $8.9$, which is much better than that of N-clustering. Whereas, for RF, the reverse trend becomes apparent, where N-clustering outperforms 1-clustering. However, in both scenarios, the differences seem almost negligibly small. As a result, k-Clustering using the POC approach is found to enhance the overall forecasting performance over the baseline approaches, which actually exhibit only slight differences in MSE.

\begin{table}[t]
\caption{MSE Loss of all the examined scenarios, 2- and 20-Clustering, as well as 1- and N-Clustering.}
\centering
\begin{tabular}{lccc}
{} &       VAR &       RF &        EBM \\
\midrule
2-Clustering (PDC)      &  9.075 & 9.024  &  8.862  \\
2-Clustering (POC)      &  8.865 & 8.823  & 8.741  \\
20-Clustering (PDC)      &  9.098 &  8.860 &  8.833 \\
20-Clustering (POC)      &  8.428 &  8.362 &  8.479 \\
1-Clustering &  9.072 &  9.089 &  8.904  \\
N-Clustering &  9.072 &   9.008 &  9.052 \\
\bottomrule
\end{tabular}
\label{tab:2}
\end{table}

\section{Discussion - Future Work}
According to the evaluation results, the superiority of the second approach, i.e. clustering based on performance, is apparent over the first one, which is based on coefficients. This was confirmed by both the Silhouette coefficients and forecasting performance. Regarding the Silhouette analysis, the fact that the produced scores did not exceed $0.17$ shows that clustering can not be characterized as meaningful. However, in real-world datasets, because of their structure complexity, values close to the maximum possible, that is 1, are not realistically expected. Thus, in this case, we should evaluate the scores in comparison to the produced values of all the rest examined experiments. Moreover, regarding performance, it is reasonable that lower errors found in the second approach, as in principles that clustering approach is optimized on the same metric, total MSE. 

According to the aforementioned evaluation measures, there was a disagreement with regards to the best chosen clustering parameters. For example, increasing the number of extracted clusters seems to give a lower error, whereas the opposite holds for the Silhouette coefficient. Thus, a trade-off analysis regarding the number of clusters is necessary to be further studied.

Another critical point of clustering was its evaluation in terms of stability.
For the majority of experiments, the estimated cluster labels of all individuals were quite different across the 10 running iterations, leading to low stability values. This means that the initialization step had a large impact on the resulting data partition, but not always observed on the test performance error. Thus, despite low stability, close values for the performance loss were found. Even, using an initialization method similar to kmeans++ only leads to local optimal solutions with respect to the defined clustering optimization. Therefore, this problem of initialization bias needs to be further investigated in more datasets.

Furthermore, all clustering experiments of the second approach, POC, compared favorably to the baseline scenarios, that are the personalized and the all-in-one group. Therefore, since potential benefits have been already found when using clustering derived-information, more advanced integrating approaches need to be studied. For example, transfer learning methods along with clustering is a way worth exploring in the future \cite{transfer, ntekouli2022using}. 

Finally, as a general remark, although using the model-based estimated parameters in a clustering task has been widely applied, in real-world problems, more complex data representations may be necessary. It is likely that high-dimensional, dynamic, and possibly noisy time-series data cannot be accurately described only with parameters derived from linear/non-linear equations. Thus, more complex feature representation should be learned using deep learning techniques \cite{deep, rnn}. Using such sequential models could also be beneficial in training, including a greater number of historical values than just one previous time-point.

\section{Conclusion}

In this paper, two model-based clustering approaches have been introduced, with a twofold goal (1) to group similar individuals using their EMA time-series data and (2) to improve the forecasting performance of all individuals. The two approaches are based on different model-derived information, one based on models’ parameters, whereas the other optimized on the total performance on a predictive task. 
According to the evaluation results, besides stability which was quite low, the remaining measures indicated that the second clustering approach (POC) outperforms the former one (PDC). The approach of k-Clustering POC was also found to enhance the overall forecasting performance, compared to the baseline approaches.
Thus, the results demonstrated that the superiority of clustering performance is not a random effect arising from the fact that a mixture of models is used. Overall, as potential benefits have been already found when using clustering derived-information, further experiments using more advanced integrating approaches need to be studied. Moreover, for a theoretical perspective, it would be interesting to produce explanations about the structures and important factors of the derived clusters, since explainable models were already used.

\subsubsection{Acknowledgements} 
This study is part of the project ``New Science of Mental Disorders" (\url{www.nsmd.eu}), supported by the Dutch Research Council and the Dutch Ministry of Education, Culture and Science (NWO gravitation grant number 024.004.016).
\bibliographystyle{plain} 
\bibliography{Bibl.bib} 
\end{document}